\documentclass[11pt,letterpaper]{article}
\usepackage{emnlp2017}
\usepackage{times}
\usepackage{latexsym}

\usepackage{CJK}  
\usepackage{url}
\usepackage{bm}
\usepackage{graphicx}
\usepackage{amsmath}
\usepackage{amssymb}
\usepackage{xcolor}
\usepackage{xspace}
\usepackage{multirow}

\newcommand{\tabincell}[2]{\begin{tabular}{@{}#1@{}}#2\end{tabular}}

\emnlpfinalcopy

\def\emnlppaperid{***}

\title{Learning to Predict Charges for Criminal Cases with Legal Basis}

\author{Bingfeng Luo$^1$, Yansong Feng$^{*1}$, Jianbo Xu$^2$, Xiang Zhang$^2$ \and Dongyan Zhao$^1$ \\
 {$^1$Institute of Computer Science and Technology, Peking University, China}\\
 {$^2$Beijing Institute of Big Data Research, China}\\
{\tt \{bf\_luo, fengyansong, xujb, xiang.zhang, zhaody\}@pku.edu.cn}\\
 }

\date{}

\begin{document}
\begin{CJK*}{UTF8}{gbsn}  

\maketitle

\begin{abstract}
The charge prediction task is to determine appropriate charges for a given case, which is helpful for legal assistant systems where the user input is fact description.
We argue that relevant law articles play an important role in this task, and therefore propose an attention-based neural network method to jointly model the charge prediction task and the relevant article extraction task in a unified framework.
The experimental results show that, besides providing legal basis, the relevant articles can also clearly improve the charge prediction results, and our full model can effectively predict appropriate charges for cases with different expression styles.

\end{abstract}
\section{Introduction}
The task of automatic charge prediction is to determine appropriate charges, 
such as \emph{fraud}, \emph{larceny} or \emph{homicide}, for a case by analyzing its textual fact description.
Such techniques are crucial for legal assistant systems, 
where users could find similar cases or possible penalties by describing a case with their own words.
This is helpful for non-legal professionals to get to know the legal basis of their interested cases, e.g., cases they or their friends are involved in,
since the massive legal materials 
and the lack of knowledge of legal jargons
make it hard for outsiders to do it on their own.



\begin{figure*}[t!]
\begin{center}
\includegraphics[width=0.97\textwidth]{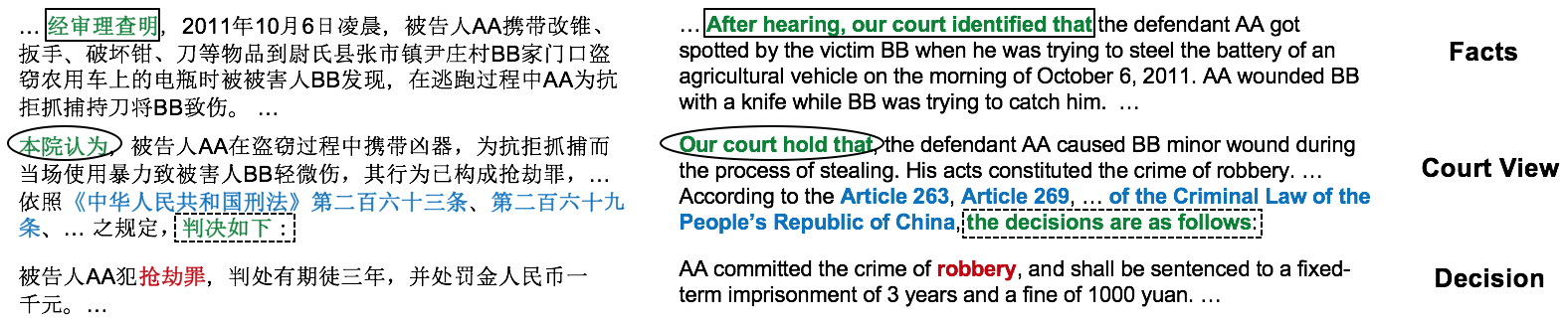}	
\caption{An example judgement document excerpt of a criminal case in our dataset. Names are anonymized as AA and BB.
Rectangulars, ellipses and dashed rectangulars refer to the clauses that usually indicate the beginning of the facts part, the court view part and the decision part, respectively. Articles and charges are extracted with regular expressions and a charge list.
}
\label{fig_example_case}
\end{center}
\vspace{-.5em}
\end{figure*}

However, predicting appropriate charges based on fact descriptions is not trivial:
(1) The differences between two charges can be subtle, for example, in the context of criminal cases in China, 
distinguishing \emph{intentional homicide} from \emph{intentional injury} would require to determine, 
from the fact description, whether the defendant  \textit{intended to kill} the victim, 
or just \textit{intended to hurt the victim, who, unfortunately died of severe injury}.
(2) Multiple crimes may be involved in a single case, which means 
we need to conduct charge prediction in the multi-label classification paradigm. 
(3) Although we can expect an off-the-shelf classification model to learn to label a case with corresponding charges 
through massive training data,  it is always more convincing to make the prediction with its involved law articles explicitly shown to the users, as legal basis to support the prediction. 
This issue is crucial in countries using \textit{the civil law system}, e.g., China (except Hong Kong), 
where judgements are made based on statutory laws only. 
For example, in Fig.~\ref{fig_example_case}, a judgement document in China always includes relevant law articles (in the court view part) to support the decision.  
Even in countries using \textit{the common law system}, e.g., the United States (except Louisiana), 
where the judgement is based mainly on decisions of previous cases, there are still some 
statutory laws that need to be followed when making decisions. 


Existing attempts formulate the task of automatic charge prediction as a single-label classification problem, 
by either adopting a k-Nearest Neighbor (KNN)~\cite{LIU2004case,liu2006exploring} as the classifier 
with shallow textual features,
or manually designing key factors for specific charges to help text understanding~\cite{lin2012exploiting},
which make those works hard to scale to more types of charges. 
There are also works addressing a related task, finding the law articles that are involved in a given case.
A simple solution is to convert  this multi-label problem 
into a multi-class classification task by only considering a fixed set of article 
combinations~\cite{liu2005classifying,liu2006exploring}, which
can only be applied to a small set of articles and does not fit to real applications.  
Recent improvement takes a two-step approach by performing 
a preliminary classification first and then re-ranking the results with word-level and article-level features~\cite{liu2015predicting}. 
These efforts  advance the applications of machine learning and natural language processing methods into legal assistance services, however, they are still in an early stage, e.g., relying on expert knowledge, 
using relatively simple classification paradigms, and shallow textual analysis. More importantly, related tasks, e.g., charge prediction and relevant article extraction,  are 
treated independently, 
ignoring the fact that they could benefit from each other.




Recent advances in neural networks enable us to jointly model charge prediction and relevant 
article extraction in a unified framework, 
where the latent correspondence from the fact description about a case to its related law articles 
and further to its charges can be explicitly addressed by a two-stack attention mechanism.
Specifically,   
we use a sentence-level and a document-level Bi-directional Gated Recurrent Units (Bi-GRU)~\cite{bahdanau2015neural} with a stack of fact-side attention components to model the correlations among words and sentences, in order to capture the whole story as well as important details of the case. 
Given the analysis of the fact description, we accordingly learn a stack of article-side attention components to attentively select the most supportive law articles from the statutory laws to support our charge prediction, which is investigated in the multi-label paradigm.

We evaluate our model in the context of  predicting charges for criminal cases in China. 
We collect publicly available judgement documents from China's government website, 
from which we can automatically extract 
\textit{fact descriptions}, \textit{relevant law articles} and \textit{the charges} using simple rules, as shown in Figure~\ref{fig_example_case}.
Experimental results show that our neural network method can effectively predict appropriate charges for a given case, and also provide relevant law articles as legal basis to support the prediction. 
Our experiments also provide quantitive analysis about the effect of fact-side and article-side information on charge predicion, 
and confirm that, apart from providing legal basis, relevant articles also contain useful information that can help to improve charge prediction in the civil law system.
We also examine our model on the news reports about criminal cases. 
Although trained on judgement documents, our model can still achieve promising performance on news data, 
showing a reasonable generalization ability over different expression styles.

\section{Related Work}
\label{sec_related_work}
The charge prediction task aims at finding appropriate charges based on the facts of a case. Previous works consider this task in a multi-class classification framework, which takes the fact description as input and outputs a charge label. 
\cite{LIU2004case,liu2006exploring} use KNN to classify 
12 and 6 criminal charges in Taiwan.
However, except for the inferior scalability of the KNN method, their word-level and phrase-level features are too shallow to capture  sufficient evidence to distinguish similar charges with subtle differences. 
\cite{lin2012exploiting} propose to make deeper understanding of a case by identifying charge-specific factors that are manually designed for 2 charges. This method also suffers from the scalability issue due to the human efforts required to design and annotate these factors for each pair of charges. Our method, however, employs Bi-GRU and a two-stack attention mechanism to make comprehensive understanding of a case without relying on explicit human annotations. 

Within the civil law system,
there are some works focusing on identifying applicable law articles for a given case. 
\cite{liu2005classifying,liu2006exploring} convert this multi-label problem into a multi-class classification problem by only considering a fixed set of article combinations, which cannot scale well since the number of possible combinations will grow exponentially when a larger set of 
law articles are considered.
\cite{liu2015predicting} instead design a scalable two-step approach 
by first using Support Vector Machine (SVM) for preliminary article classification, and then 
re-ranking the results using 
word level features and co-occurence tendency among articles.
We also use SVM to extract top $k$ candidate articles, but further adopt Bi-GRU and article-side attention to better understand the texts and exploring the correlation among articles. 
More importantly, we optimize the article extraction task within our charge prediction framework,  which not only provides another view to understand the facts, but also serves as legal basis to support the final decision.

Another related thread of work 
is to predict the overall outcome of a case. The target can be 
which party will the outcome side with~\cite{aletras2016predicting}, or whether the present court will affirm or reverse the decision of a lower court~\cite{katz2016general}. Our work differs from them in that, instead of binary outcome (the latter one also contains an \emph{other} class), we step further to focus on the detailed results of a case, i.e., the charges, where the output may contain multiple labels. 


We also share similar spirit with the legal question answering task~\cite{COLIEE14}, which aims at answering the yes/no questions in the Japanese legal bar exams, that we all believe that relevant law articles are important for decisions in the civil law system. 
Different from ours, this task requires participants to extract relevant Japanese Civil Code articles first, 
and then use them to answer the yes/no questions. 
The former phase is often treated as an information retrieval task, and the latter phase is considered as a textual entailment task~\cite{kim2014legal,carvalho2016lexical}. 

In the field of artificial intelligence and law, there are also works trying to find relevant cases given the input query~\cite{raghav2016analyzing,chen2013text}, which is crutial for decision making in the common law system. 
Rather than finding relevant cases, our work focuses on predicting specific charges, and we also emphasize the importance of law articles in decision making,
which is important in the civil law system where the decisions are made based solely on statutory laws.

Our work is also related to the task of document classification, but mainly differs in that we also need to automatically identify applicable law articles to support and improve the charge prediction.
Recently, various neural network (NN) architectures such as Convolutional Neural Network (CNN)~\cite{kim2014convolutional} and Recurrent Neural Network (RNN) have been used for document embedding, which is further used for classification.
\cite{tang2015document} propose a two-layer scheme, RNN or CNN for sentence embedding, and another RNN for document embedding.
\cite{yang2016hierarchical} further use global context vectors to attentively distinguish informative words or sentences from non-informative ones during embedding, which we share similar spirit with. 
But, we take a more flexible and descriptive two-stack attention mechanism, one stack for fact embedding, and the other for article embedding which is  dynamically generated for each instance according to the fact-side clues as extra guidance.
%
Another difference is the multi-label nature of our task, where, rather than optimizing as multiple binary classification tasks~\cite{nam2014large}, 
we convert the multi-label target to label distribution during training with cross entropy as loss function~\cite{kurata2016improved}, and use a threshold tuned on validation set to produce the final prediction, which performs better in our pilot experiments.
\section{Data Preparation} 
Our data are collected from China Judgements Online\footnote{\url{http://wenshu.court.gov.cn}}, where the Chinese government has been publishing judgement documents since 2013.
We randomly choose 50,000 documents for training, 5,000 for validation and 5,000 for testing. To ensure enough training data for each charge, we only classify the charges that appear more than 80 times in the training data, and treat documents with other charges as negative data. As for law articles, we consider those in the Criminal Law of the People's Republic of China. The resulting dataset contains 50 distinct charges, 321 distinct articles, averagely 383 words per fact description, 3.81 articles per case, and 3.56\% cases with more than one charges.

An example judgement document is shown in Figure \ref{fig_example_case}, where we highlight the indicator clauses that we used to  
divide a document into three pieces 
and extract \textit{fact description}, \textit{articles}, and \textit{charges} from each piece, respectively.
We use a manually collected charge list to identify all the charges, and law articles are extracted by regular expressions\footnote{The regular expression used to extract law articles: ``第[、零〇一二两三四五六七八九十百千0-9]+条(之[一二两三四五六七八九十])?)''}.
The extracted charges and articles are considered as gold standard charges and articles for the corresponding fact description.
We also masked all the charges in fact descriptions, since although rare, charge names sometimes may appear in the fact description part.


Currently, it is hard and expensive to match the facts related to different defendants with their corresponding charges. We therefore only consider the cases with one defendant, and leave the challenging multi-defendant cases for future work. Although this simplification may change the real-world charge distribution, it enables us to automatically build large scale high quality dataset without relying on annotations from legal practitioners.

\section{Our Approach}
\begin{figure}[t!]
\begin{center}
\includegraphics[width=0.48\textwidth]{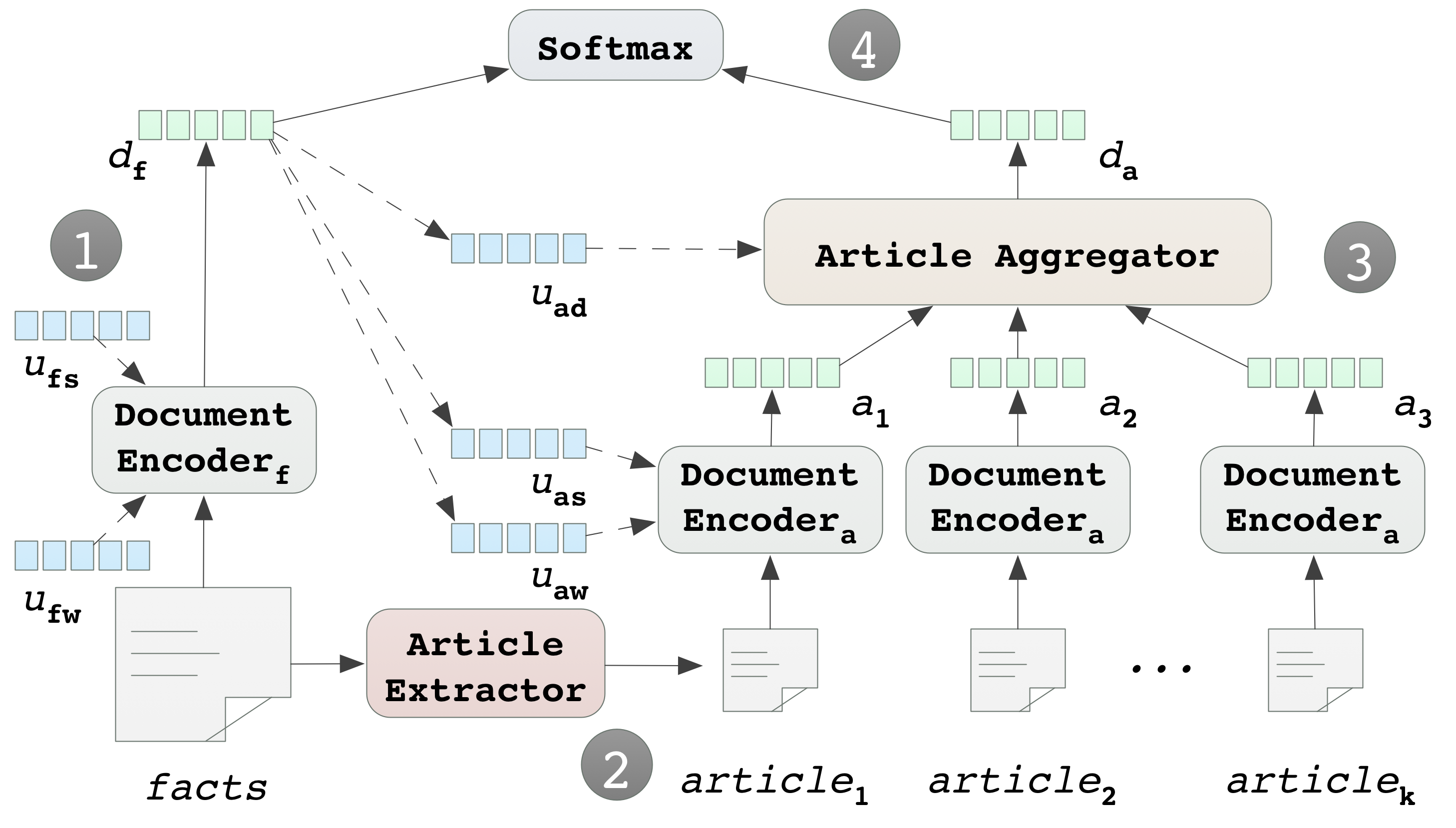}	
\caption{Overview of Our Model}
\label{fig_model_framework}
\end{center}
\vspace{-1em}
\end{figure}

As depicted in Fig.~\ref{fig_model_framework}, our approach contains the following steps:
(1) The input fact description is fed to a document encoder to generate the fact embedding $\mathbf{d}_f$, where $\mathbf{u}_{fw}$ and $\mathbf{u}_{fs}$ are global word-level and sentence-level context vectors used to attentively select informative words and sentences.
(2) Concurrently, the fact description is also passed to an article extractor to find top $k$ relevant law articles. 
(3) These articles are embedded by another document encoder, and  passed to an article aggregator to attentively select supportive articles, and produce the aggregated article embedding $\mathbf{d}_a$. Specifically, three context vectors, i.e., $\mathbf{u}_{aw}$, $\mathbf{u}_{as}$ and $\mathbf{u}_{ad}$, are dynamically generated from $\mathbf{d}_f$, to 
produce attention values within
the article document encoder and the article aggregator. 
(4) Finally, $\mathbf{d}_f$ and $\mathbf{d}_a$ are concatenated and passed to a softmax classifier to predict the charge distribution for the input case.

\subsection{Document Encoder}
\label{sec_doc_encoder}
\begin{figure}[htbp]
\vspace{-.5em}
\begin{center}
\includegraphics[width=0.4\textwidth]{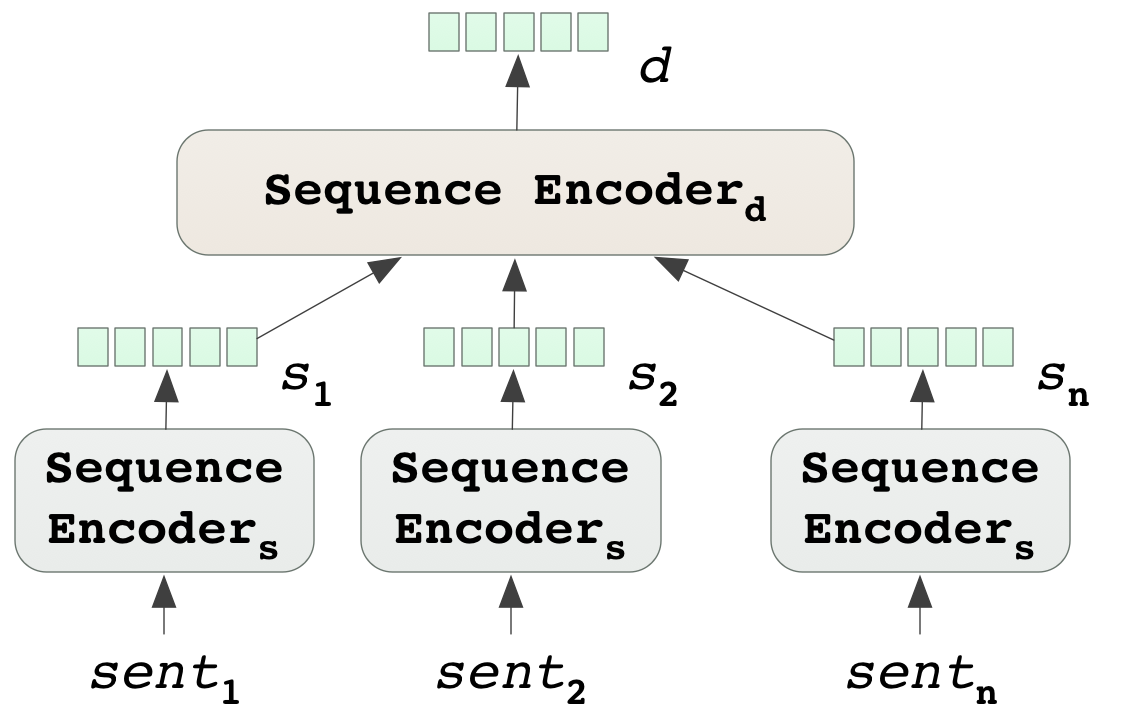}	
\caption{Document Encoder Framework}
\label{fig_doc_encoder}
\end{center}
\end{figure}

Intuitively, a sentence is a sequence of words and a document is a sequence of sentences. 
The document embedding problem, therefore, can be converted to two sequence embedding problems \cite{tang2015document,yang2016hierarchical}. As shown in Fig.~\ref{fig_doc_encoder}, we can first embed each 
sentence using a sentence-level sequence encoder, and then aggregate them with a document-level sequence 
encoder to produce the document embedding~$\mathbf{d}$. 
While these two encoders can have different architectures, we use the same here for simplicity.

\begin{figure}[htbp]
\vspace{-1em}
\begin{center}
\includegraphics[width=0.45\textwidth]{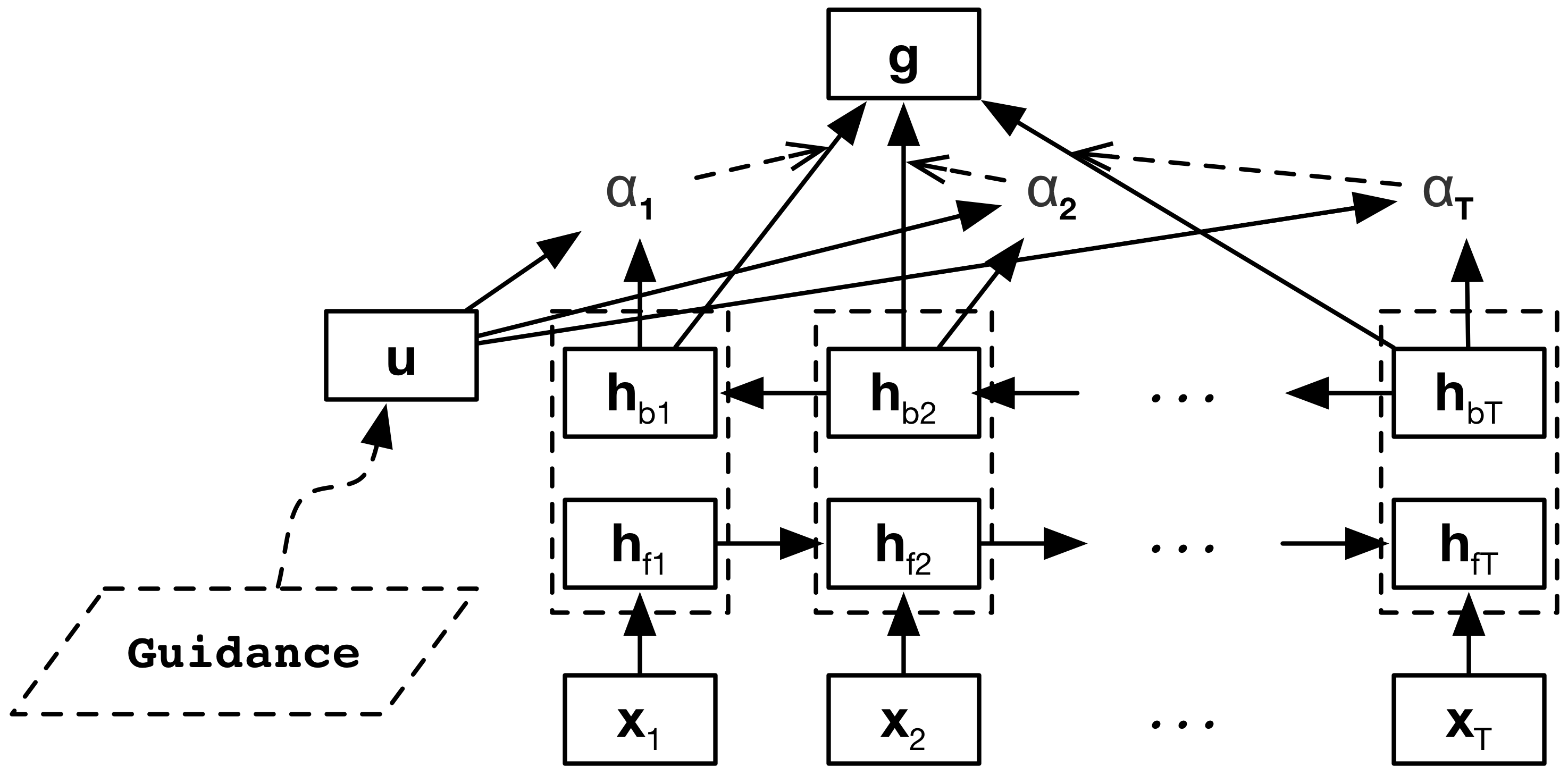}	
\caption{Attentive Sequence Encoder}
\label{fig_seq_encoder}
\end{center}
\vspace{-1em}
\end{figure}

\paragraph{Bi-GRU Sequence Encoder} 
A challenge in building a sequence encoder is how to take the correlation among different elements into consideration. A promising solution is Bi-directional Gated Recurrent Units (Bi-GRU)~\cite{bahdanau2015neural}, which encodes the context of each element by using a gating mechanism to track the state of sequence.
Specifically, Bi-GRU first uses a forward and a backward GRU~\cite{cho2014learning}, which is a kind of RNN, to encode the sequence in two opposite directions, and then concatenates the states of both GRUs to form its own states. 

Given a sequence $[\mathbf{x}_1, \mathbf{x}_2, ..., \mathbf{x}_T]$ where $\mathbf{x}_t$ is the input embedding of element $t$, the state of Bi-GRU at position $t$ is:
\begin{equation}
\mathbf{h}_t = [\mathbf{h}_{ft}, \mathbf{h}_{bt}]
\end{equation}
where $\mathbf{h}_{ft}$ and $\mathbf{h}_{bt}$ are the states of the forward and backward GRU at position $t$. 
The final sequence embedding is either the concatenation of $\mathbf{h}_{fT}$ and $\mathbf{h}_{b1}$
or simply the average of $\mathbf{h}_t$.





\paragraph{Attentive Sequence Encoder}
\label{sec_att_seq_encoder}
However, directly using $[\mathbf{h}_{fT}, \mathbf{h}_{b1}]$ for sequence encoding
often fails to capture all the 
information when the sequence is long, while using the average of $\mathbf{h}_t$ also has the drawback of treating useless elements equally with informative ones. 
Inspired by \cite{yang2016hierarchical}, we use a context vector to attentively aggregate the elements, but instead of using a global context vector, we allow the context vector to be dynamically generated when extra guidance is available (see Sec.~\ref{sec_article_encoder}).

As shown in Fig.~\ref{fig_seq_encoder}, given the Bi-GRU state sequence $[\mathbf{h}_1, \mathbf{h}_2, ..., \mathbf{h}_T]$, 
our attentive sequence encoder calculates a sequence of attention values $[\alpha_1, \alpha_2, ..., \alpha_T]$, 
where $\alpha_t \in [0, 1]$ and $\sum_t{\alpha_t}=1$. The final sequence embedding $\mathbf{g}$ is calculated by:
%
\begin{equation}
\mathbf{g} = \sum_{t=1}^{T}{\alpha_t \mathbf{h}_t};\ \ \ 
\alpha_t=\frac{exp(tanh(\mathbf{W} \mathbf{h}_t)^T \mathbf{u})}{\sum_t{exp(tanh(\mathbf{W} \mathbf{h}_t)^T \mathbf{u})}}
\label{seq_embed}
\end{equation}
where $\mathbf{W}$ is a weight matrix, and $\mathbf{u}$ is the context vector to distinguish informative elements from non-informative ones. 

By using this sequence encoder for fact embedding, the fact-side attention module actually contains two components, i.e., the word-level and sentence-level, using $\mathbf{u}_{fw}$ and $\mathbf{u}_{fs}$ as their global context vectors, respectively.

\subsection{Using Law Articles} 
One of the challenges of using law articles to support charge prediction lies in the fact that statutory laws contain a large number of articles, which makes applying complex models to these articles directly time-consuming, and thus hard to scale.
The multi-label nature of relevant article extraction also requires a model that can output multiple articles.
%
We thus adopt a two-step approach, 
specifically, we first build a fast and easy-to-scale classifier to filter out a large fraction of irrelevant articles, and retain the top $k$ articles. Then, we use neural networks to make comprehensive understanding of the top $k$ articles, and further use the article-side attention module to select the most supportive ones for charge prediction.

\paragraph{Top $k$ Article Extractor}
\label{sec_article_extractor}
We treat the relevant article extraction task as multiple binary classifications. Specifically, we build a binary classifier for each article, focusing on its relevance to the input case, 
which results
in 321 binary classifiers corresponding to the 321 distinct law articles 
appearing in our dataset.
When more articles are considered,
we can simply add more binary classifiers accordingly, with the existing classifiers untouched.

Similar to the preliminary classification phase of \cite{liu2015predicting},
we also use word-based SVM as our binary classifier, which is fast and performs well in text classification~\cite{joachims2002learning,wang2012baselines}. Specifically, we use bag-of-words TF-IDF features, chi-square for feature selection and linear kernel for binary classification. 

\paragraph{Article Encoder}
\label{sec_article_encoder}
Since each law article may contain multiple sentences, 
as shown in Fig.~\ref{fig_model_framework}, we also use the document encoder described in Sec.~\ref{sec_doc_encoder} to produce an embedding $\mathbf{a}_j, j\in[1, k]$, for each article in the top $k$ extracted articles. 
While using similar architecture, this article encoder differs from the fact encoder that, instead of using global context vectors, its word-level context vector $\mathbf{u}_{aw}$ and sentence-level context vector $\mathbf{u}_{as}$ are dynamically generated  for each case according to its corresponding fact embedding $\mathbf{d}_f$:
\begin{equation}
\mathbf{u}_{aw} = \mathbf{W}_w \mathbf{d}_f + \mathbf{b}_w;\ \ \ \ \mathbf{u}_{as} = \mathbf{W}_s \mathbf{d}_f + \mathbf{b}_s
\label{eq_dynamic_context_vec}
\end{equation}
where $\mathbf{W_*}$ is the weight matrix and $\mathbf{b_*}$ is the bias. 
The context vectors, $\mathbf{u}_{aw}$ and $\mathbf{u}_{as}$, are used to produce the word-level and sentence-level attention values,  respectively.
Through the dynamic context vectors, the fact embedding $\mathbf{d}_f$ actually guides our model to attend to informative words or sentences with respect to the facts of each case, rather than just selecting generally informative ones.

\paragraph{Attentive Article Aggregator}
The article aggregator aims to find supportive articles for charge prediction from the top $k$ extractions, and accordingly produce an aggregated article embedding.
%
%
%
Although the order of the top $k$ extracted articles is not fully reliable, \cite{vinyals2016matching} suggests that it is still beneficial to use a bi-directional RNN to embed the context of each element even in a set, where the order does not exist.
In our task, bi-directional RNN can help to utilize the co-occurrence tendency of relevant articles.

Specifically, we use the attentive sequence encoder in Sec.~\ref{sec_att_seq_encoder} 
to produce the aggregated article embedding $\mathbf{d}_a$. 
Again, to guide the attention with fact descriptions, we dynamically generate the article-level context vector $\mathbf{u}_{ad}$ by:
\begin{equation}
\mathbf{u}_{ad} = \mathbf{W}_d \mathbf{d}_f + \mathbf{b}_d
\end{equation}

The attention values produced by the attentive sequence encoder can be seen as the relevance of each article to the input case, which can be used to rank and filter the top $k$ articles. The results can be shown to users as legal basis for charge prediction.

\subsection{The Output}
To make the final charge prediction, we first concatenate the document embedding $\mathbf{d}_f$ and the aggregated article embedding $\mathbf{d}_a$, and feed them to two consecutive full connection layers to generate a new vector $\mathbf{d}'$, which is then passed to a softmax classifier to produce the predicted charge distribution. We use the validation set to determine a threshold $\tau$, and consider all the charges with output probability higher than $\tau$ as positive predictions.
The input to the first full connection layer can also be only $\mathbf{d}_f$ or $\mathbf{d}_a$, which means we use only fact or article to make the prediction.

The loss function for training is cross entropy:
\begin{equation}
\label{original_loss}
Loss= -\sum_{i=1}^N\sum_{l=1}^L{y_{il} log(o_{il})}
\end{equation} 
where $N$ is the number of training data, $L$ is the number of charges, $y_{il}$ and $o_{il}$ are the target and predicted probability of charge $l$ for case $i$. The target charge distribution $\mathbf{y}_i$ is produced by setting positive labels to $\frac{1}{m_i}$ and negative ones to $0$, where $m_i$ is the number of positive labels in case $i$.

\paragraph{Supervised Article Attention}
We can also utilize the gold-standard law articles naturally in the judgement documents
to supervise the article attention during training. Specifically, given the top $k$ articles, we want the article attention distribution $\bm{\alpha}\in\mathbb{R}^k$ to simulate the target article distribution $\mathbf{t}\in\mathbb{R}^k$, where $t_j=\frac{1}{k'}$ if article $j$ belongs to the gold-standard articles and $t_j=0$ otherwise. Here $k'$ is the number of gold-standard articles in the top $k$ extractions.
%
%
We, again, use cross entropy, and the loss function is:
\begin{equation}
\label{final_loss}
Loss = -\sum_{i=1}^N(\sum_{l=1}^L{y_{il} log(o_{il})} + \beta \sum_{j=1}^k{t_{ij} log(\alpha_{ij})})
\end{equation}
where $\beta$ is the weight of the article attention loss.




\section{Experiments}
\subsection{Experimental Setup}
We use HanLP\footnote{\url{https://github.com/hankcs/HanLP}} for Chinese word segmentation and POS tagging.
Word embeddings are trained using word2vec~\cite{mikolov2013distributed} on judgement documents, web pages from several legal forums and Baidu Encyclopedia. The resulting word embeddings contain 573,353 words, with 100 dimension.
We randomly initialize a 50-d vector for each POS tag, which is concatenated with the word embedding as the final input.
Each GRU in the Bi-GRU is of size 75, the two full connection layers are of size 200 and 150.
The relevant article extractor generates top 20 articles, the weight of the article attention loss ($\beta$ in Eq.~\ref{final_loss}) is 0.1, and prediction threshold $\tau$ is 0.4.
We use Stochastic Gradient Descent (SGD) for training, with learning rate 0.1, and batch size 8.

We compare our full model with two variations: without article attention supervision and only using facts for charge prediction. 
The latter one is similar to the state-of-art document classification model~\cite{yang2016hierarchical}, but adapted to the multi-label nature of our problem.
We also implement an SVM model, 
which is effective and scales well in many fact-description-related tasks in the field of artificial intelligence and law~\cite{liu2015predicting,aletras2016predicting}.
Specifically, the SVM model takes bag-of-words TF-IDF features as input, and uses chi-square to select top 2,000 features.

\begin{table}
\setlength{\tabcolsep}{0.23em}
\centering
\small{
\begin{tabular}{|c|c|c|c|}
\hline
\multirow{2}{*}{\textbf{Model}}				& \tabincell{c}{\textbf{Precision}} 	& \tabincell{c}{\textbf{Recall}} 		& \tabincell{c}{\textbf{F1}} 	\\
\cline{2-4}
                                               & \multicolumn{3}{c|}{\tabincell{c}{ (\textit{Micro-/Macro-}) }}\\
\hline
\textit{SVM\_fact} 				& \textbf{93.94}/79.53					& 77.66/49.54  					& 85.03/61.05 				 	\\
\hline
\textit{SVM\_art} 			& 82.12/42.90					& 61.23/39.56  					& 70.15/41.16				 	\\
\hline
\textit{SVM\_fact\_art} 			& 91.77/71.33					& 72.10/45.85  					& 80.76/55.82				 	\\
\hline
\textit{NN\_fact}				& 91.30/\textbf{83.32}			& 87.39/74.99  					& 89.31/78.94					\\
\hline
\textit{NN\_art} 			& 90.09/81.50				& 86.10/69.62				& 88.05/75.10		\\
\hline
\textit{NN\_fact\_art}			& 90.79/83.07					& 88.42/75.73  					& 89.59/79.23					\\
\hline
\textbf{\textit{NN\_fact\_supv\_art}} 	& 91.80/82.44 					& \textbf{88.67/78.62} 			& \textbf{90.21/80.48} 		 	\\
\hline
\hline
\textit{SVM\_fact\_gold\_art} 	& \textbf{98.97}/94.58			& 95.39/83.21  					& 97.15/88.53					\\
\hline
\textbf{\textit{NN\_fact\_gold\_art}} 		& 98.78/\textbf{95.26} 			& \textbf{98.24/95.57} 			& \textbf{98.51/95.42} 			\\
\hline
\end{tabular}
}
\vspace{-.5em}
\caption{Charge prediction results. 
Left and right side of the slash refer to micro and macro statistics, respectively.
\texttt{gold\_art} refers to using gold standard articles mentioned in judgements (marked in blue in Fig. 1), which is the upper bound for article-related modules.}
\label{tabble_main_results}
\vspace{-.5em}
\end{table}


\subsection{Charge Prediction Results}
\label{sec_main_results}
The charge distribution is imbalanced, and the top 5 charges take more than 60\% of the cases.
Therefore, we evaluate the charge prediction task using precision, recall and F1, in both micro- and macro-level.
The macro-precision/recall are calculated by 
averaging the precision and recall of
each charge, and the micro-precision/recall are averaged over each prediction. 




As shown in Table~\ref{tabble_main_results}, 
the basic \texttt{SVM\_fact} model,
which only takes fact descriptions as input, indeed proves to be a strong baseline.
By contrast, our corresponding neural network model (\texttt{NN\_fact}), which also only uses facts for prediction, outperforms \texttt{SVM\_fact} 
by about 4\% in micro-F1.
Since \texttt{NN\_fact} benefits from the pre-trained word embeddings,
the two-level Bi-GRU architecture, and the fact-side attention module,
it can attentively recognize informative expressions from the description and 
better capture the underlying correspondence from fact descriptions to appropriate charges,
even when there is less overlap in the words used among cases with the same charge,
or when there are limited data (i.e., infrequent charges). This may explain that NN models 
have more balanced performance over different charges, leading to more prominent improvements over 
SVM ones in macro metrics, which usually have a strong bias towards frequent charges.
%


When we use both facts and extracted relevant law articles (that are admittedly noisy),
the SVM version (\texttt{SVM\_fact\_art}) drops by around~5\% than \texttt{SVM\_fact}, showing that the SVM model 
cannot benefit from the extracted, thus noisy, relevant articles in such a straightforward way.
However, our NN version (\texttt{NN\_fact\_article}) can still learn from  the noisy article extractions through 
attentively aggregating those extracted articles even without direct guidance,
thus improves \texttt{NN\_fact} by around 0.4\%.
Furthermore, if we use the gold standard articles during training as supervision for the 
article attention (our full model, \texttt{NN\_fact\_supv\_art}), the results can be further improved, achieving 
90.21\% and 80.48\% in micro- and macro-F1, respectively.
The improvements made by using relevant law articles actually indicates the nature of the civil law system that judgements are made based on statutory laws.

However, if we only use the extracted relevant articles to make prediction (\texttt{SVM\_art} and \texttt{NN\_art}\footnote{
\texttt{NN\_art} uses fact embeddings to attentively aggregate relevant articles, but only use 
the aggregated article embedding $\mathbf{d}_a$, without fact embedding  $\mathbf{d}_f$,  for charge prediction.}),
the performance becomes worse.
Even with the proved-helpful attentive aggregator, the model performs worst among all NN variants (though still better than \texttt{SVM\_fact}). 
This indicates that it is necessary to consider both facts and relevant law articles for charge prediction, and,
the fact that \texttt{NN\_fact} outperforms \texttt{NN\_art} also indicates that although the judgments are made based on the statutory laws in the civil law system, the logic employed by the court when making decisions, to some extent, may be implicitly captured through massive fact-charges paris.

Now the question is: \textit{how much improvement can we have 
if we can make full use of
the relevant law articles within the civil law system?}
Let us consider an ideal situation where we can access both fact descriptions and gold standard law articles during testing, which could be considered as an upper bound scenario.
The SVM version (\texttt{SVM\_fact\_gold\_art}) significantly outperforms 
\texttt{SVM\_fact\_art} by more than 30\% in macro-F1.
And the NN version 
(\texttt{NN\_fact\_gold\_art}) outperforms 
\texttt{NN\_fact\_supv\_art} by over 8\%.
%
These comparisons confirm again that 
law articles play an important role for automatic judgement prediction, but 
the extracted relevant articles inevitably contain noise, which should be properly 
handled, e.g., using an attentively aggregation mechanism to distill 
valuable evidence to support charge prediction.

\paragraph{Case Study}
We study the model outputs and find certain star-like confusion patterns among the charges. For example, \emph{intentional injury} 
is often confused with multiple charges like 
\emph{intentional homicide} (when the victim is dead, the difference is 
 whether the defendant intends to kill  or just hurt the victim) and \emph{picking quarrels and provoking troubles}
(there may also exist injuries here).
These charges usually share some similar fact descriptions, e.g., how the injuries are caused, and since \emph{intentional injury} appears more frequently than the others, \texttt{SVM\_fact} thus outputs \emph{intentional injury} in most situations, and fails to distinguish these charges.
However, by using Bi-GRU and the attention mechanism, \texttt{NN\_fact} can attend to important details of the facts and significantly improves the performance on these charges.
When the direct supervision for articles is available,  \texttt{NN\_fact\_supv\_art}  can enhance the interaction between certain pairs of fact descriptions and law articles, which helps to capture the subtle differences among similar charges, and further improves the performance on these situations.


\subsection{Article Extraction Results}
We also evaluate our SVM article extractor, which achieves 
77.60\%, 88.96\%, 94.21\% and 96.53\% recall regarding the top 5, 10, 20 and 30 articles, respectively.
Although simple, the SVM extractor can obtain over 94\% recall for top 20 articles, which is good enough for further refinement.
However, the micro-F1 of the extractor is only 61.08\% in the test set, 
which will lead to severe error propagation problem 
if we use the prediction results of the extractor directly.
Therefore, we design the article attention mechanism to handle the noise in the top 20 articles.



\begin{table}
\centering
\small{
\begin{tabular}{|c|c|c|c|c|}
\hline
\bm{$\beta$}							& \textbf{Prec@1} 		& \textbf{MAP} 			& \textbf{Charge\_F1} \\
\hline
\textit{0} 								& 60.94								& 61.61 						& 89.59/79.23 	\\
\hline
\textit{0.01} 						& 81.06								& 78.00							& 89.77/79.48 	\\
\hline
\textit{0.1} 							& 87.90								& 83.39							& \textbf{90.21/80.48} 	\\
\hline
\textit{0.5} 							& 91.44								& 86.95							& 89.93/79.67 	\\
\hline
\textit{1} 								& \textbf{92.66}			& \textbf{88.24}		& 89.83/78.66 	\\
\hline
\end{tabular}
}
\caption{Refined Article Extraction Performance}
\label{tab_article_att}
\end{table}


Table \ref{tab_article_att} shows the re-ranking results of our article attention module   (column 2-3) 
and the corresponding charge prediction performances (column 4), under different weights for article attention ($\beta$ in Eq.~\ref{final_loss}). \texttt{Prec@1} refers to top 1 precision, and \texttt{MAP} refers to mean average precision.
We can see that, even if there is no supervision over the article attention ($\beta=0$), our model still has reasonable performance on re-ranking the $k$ articles. When the attention supervision is employed, the extraction quality improves significantly, and keeps increasing as $\beta$ goes up.
However, the charge prediction performance does not always increase with the article extraction quality, and the best performance is achieved when $\beta=0.1$. This is not surprising, since  there exists a tradeoff between the benefits of more accurate article extraction and the less model capacity left for charge classification due to the increased emphasis on the article extraction performance. The promising article extraction results also confirm the ability of our model to provide legal basis for the charge prediction.

\begin{table}
\centering
\small{
\begin{tabular}{|c|c|c|c|}
\hline
\textbf{Model}												& \textbf{Precision} 				& \textbf{Recall} 				& \textbf{F1} 	\\
\hline
\textit{SVM\_fact} 													& \textbf{100.00}						& 40.20  									& 57.34 				 	\\
\hline
\textit{NN\_fact}														& 87.14											& 59.80 									& 70.93					\\
\hline
\textit{NN\_fact\_art}					& 87.18											& 66.67 									& 75.56					\\
\hline
\textbf{\textit{NN\_fact\_supv\_art}} 	& 90.00 										& \textbf{70.59} 					& \textbf{79.12} 		 	\\
\hline
\end{tabular}
}
\caption{Performance (micro statistics) on News}
\label{tabble_news_results}
\vspace{-.5em}
\end{table}

\subsection{Performance on News Data}
There are usually clear differences between the expressions used by legal practitioners and people without legal background,
thus it is important to see how our model will perform on fact descriptions written by non-legal professionals.

We create a news dataset by asking 3 law school students to annotate the appropriate charges for 100 social news reports about criminal cases
from two news websites\footnote{\url{http://news.cn} and \url{http://people.com.cn}},
with 262 words on average and 25 distinct charges. 
The $\kappa$ value is 0.83, indicating good consistency. 
The annotators are asked to have a disscussion to achieve an aggreement on inconsistent annotations.
The results are shown in Table \ref{tabble_news_results}, where we only report micro statistics due to 
the relatively small size of the dataset compared with the number of distinct charges.


We can see that, \texttt{SVM\_fact} suffers a significant drop in F1 on the news data, 
confirming the gap between the expressions used by legal practitioners and non-legal professionals,
given the BOW nature of \texttt{SVM\_fact}.
Although \texttt{SVM\_fact} cannot generalize well, the patterns learned by \texttt{SVM\_fact} are reliable in themselves, leading to a high precision.
It is not surprising that our NN models also suffer from the expression differences, 
but due to the effectiveness of our NN architecture, with about 10\%$\sim$15\% less absolute drop in F1, and \texttt{NN\_fact\_supv\_art} can still achieve 79.12\% in F1. 
For example, the word 暴打$\ $(beat up) is seldom used in judgement documents,
making it hard for \texttt{SVM\_fact} to correctly utilize  暴打$\ $as an indicator for injury related charges, but, 
our NN models can associate it with its  near-synonymy  殴打$\ $(hit), which is a formal expression in judgement documents.
Furthermore, the clear improvements from \texttt{NN\_fact} to \texttt{NN\_fact\_art}, and further to \texttt{NN\_fact\_supv\_art} prove again the importance of relevant law articles 
in supporting the charge prediction, even in news domain.
The reasonable performance on news data also shows that our method do have the ability to help non-legal professionals.

\section{Conclusion}
In this paper, we propose an attention-based neural network framework that can jointly model the charge prediction task and the relevant article extraction task, where the weighted relevant articles can serve as legal basis to support the charge prediction.
The experimental results on judgement documents of criminal cases in China show the effectiveness of our model on both charge prediction and relevant article extraction.
The comparison of different variants of our model also indicates the importance of law articles in making judicial decisions in the civil law system.
By experimenting on news data, we show that, although trained on judgement documents, our model also has reasonable generalization ability on fact descriptions written by non-legal professionals.
While promising, our model still cannot explicitly handle multi-defendant cases, and there is also a clear gap between our model and the upper bound improvement that relevant articles can achieve. We will leave these challenges for future work.

\section*{Acknowledgement}
This work is supported by the National High Technology R\&D Program of China (2015AA015403); 
the National Natural Science Foundation of China (61672057, 61672058);
KLSTSPI Key Lab. of Intelligent Press Media Technology.

\bibliography{emnlp2017}
\bibliographystyle{emnlp_natbib}

\end{CJK*}  
\end{document}